\documentclass[conference]{IEEEtran}
\IEEEoverridecommandlockouts
\usepackage{cite}
\usepackage{amsmath,amssymb,amsfonts}
\usepackage{algorithmic}
\usepackage{graphicx}
\usepackage{textcomp}
\usepackage{xcolor}
\usepackage{url}
\usepackage[nameinlink,capitalize]{cleveref}
\usepackage[font=footnotesize]{subcaption}
\usepackage{siunitx}
\usepackage{bm}
\usepackage{svg}

\usepackage{todonotes}
\def\BibTeX{{\rm B\kern-.05em{\sc i\kern-.025em b}\kern-.08em
    T\kern-.1667em\lower.7ex\hbox{E}\kern-.125emX}}
\begin{document}

\title{Online Performance Assessment of Multi-Source-Localization for Autonomous Driving Systems Using Subjective Logic
}

\author{
    \IEEEauthorblockN{
        Stefan Orf\IEEEauthorrefmark{2}\IEEEauthorrefmark{1},
        Sven Ochs\IEEEauthorrefmark{2}\IEEEauthorrefmark{1},
        Marc Ren\'{e} Zofka\IEEEauthorrefmark{2},
        J. Marius Zöllner\IEEEauthorrefmark{2}\IEEEauthorrefmark{3}
    }
    \thanks{
        \IEEEauthorrefmark{1}Equally contributing authors
    }
    \thanks{
        \IEEEauthorrefmark{2}FZI Research Center for Information Technology, 76131 Karlsruhe, Germany \texttt{\{orf, ochs, zofka, zoellner\}@fzi.de}
    }
    \thanks{
        \IEEEauthorrefmark{3}Applied Technical-Cognitive Systems, Karlsruhe Institute of Technology, 76131 Karlsruhe, Germany \texttt{marius.zoellner@kit.edu}
    }
}

\maketitle

\begin{abstract}
Autonomous driving (AD) relies heavily on high precision localization as a crucial part of all driving related software components. The precise positioning is necessary for the utilization of high-definition maps, prediction of other road participants and the controlling of the vehicle itself. Due to this reason, the localization is absolutely safety relevant. Typical errors of the localization systems, which are long term drifts, jumps and false localization, that must be detected to enhance safety. An online assessment and evaluation of the current localization performance is a challenging task, which is usually done by Kalman filtering for single localization systems. 
Current autonomous vehicles cope with these challenges by fusing multiple individual localization methods into an overall state estimation. Such approaches need expert knowledge for a competitive performance in challenging environments. This expert knowledge is based on the trust and the prioritization of distinct localization methods in respect to the current situation and environment.

This work presents a novel online performance assessment technique of multiple localization systems by using subjective logic (SL). In our research vehicles, three different systems for localization are available, namely odometry-, Simultaneous Localization And Mapping (SLAM)- and Global Navigation Satellite System (GNSS)-based. Our performance assessment models the behavior of these three localization systems individually and puts them into reference of each other. 
The experiments were carried out using the CoCar NextGen, which is based on an Audi A6. The vehicle’s localization system was evaluated under challenging conditions, specifically within a tunnel environment. The overall evaluation shows the feasibility of our approach.
\end{abstract}


\section{Introduction}
AD technology has made significant strides in recent years, with increasing levels of autonomy being deployed in both research and commercial settings. One of the most critical components enabling this progress is high-precision localization—the vehicle’s ability to determine its exact position and orientation within its environment. It is necessary for tasks such as utilizing high-definition maps, predicting the behavior of other road users, executing safe path planning, and controlling vehicle motion. As such, localization directly impacts the safety and reliability of the overall system.

Despite the maturity of localization technologies, ensuring consistently accurate positioning under real-world conditions remains a challenging task. Autonomous vehicles operate across a wide range of environments, including urban canyons, tunnels, rural roads, and GPS-denied areas, each of which can degrade the performance of certain localization methods. Typical issues include long-term drifts in position estimates, sudden jumps due to sensor anomalies, or complete failures resulting in false localization. These errors can propagate throughout the vehicle's software stack, leading to unsafe behavior if not detected in real time.

Traditional approaches, such as Kalman filtering, can estimate uncertainty in individual localization systems but are limited in their ability to evaluate the overall reliability when multiple sources—such as odometry, GNSS, and SLAM—are fused. Autonomous vehicles often use expert-tuned fusion schemes to handle diverse conditions, relying on prior knowledge to prioritize certain localization methods based on the scenario. This approach lacks adaptability and may not generalize well across environments.

In this work, we present an online performance assessment framework based on SL to evaluate and fuse the outputs of multiple localization systems. Our method quantifies the trustworthiness of each system individually and combines these into a unified performance estimate, allowing for adaptive, context-aware localization evaluation.

\section{Related Works}

Viana et al. \cite{viana_reconfigurable_2022}  propose a reconfigurable localization framework that employs a hierarchical structure with three accuracy-ordered levels, an error-detection module, and a decision block for reconfiguration. Level 3 provides the highest accuracy using LiDAR, a monocular camera, an IMU, and odometry for relative positioning. Level 2 enhances previous estimations by integrating a digital map for road-matching, while Level 1 relies on GPS/INS-based localization. The error-detection block continuously assesses localization accuracy by utilizing the approach \cite{delmotte_detection_2008}, identifying failures and updating an error-state variable, which is then processed by the decision block to determine the appropriate localization level. This ensures adaptive system performance, with evaluation only conducted in CARLA \cite{dosovitskiy_carla_2017}.

Rodríguez-Arozamena et al.  \cite{rodriguez-arozamena_fail-safe_2023} introduces a fail-safe decision architecture. The architecture inherits three distinct modules: Fallback Positioning System,  Safety Mode Decision Module, and Fail-Safe Trajectory Planning. The Safety Mode Decision has, at its core, a fuzzy logic-based decision system. The output of the fuzzy system gives three options, which correspond with each mode of operation after assessing the risk evaluation: Normal condition, Fail-Safe Operation, and Total Failure Automation Disengagement. The severity is defined by expert knowledge, and the Fail-Safe Trajectory Planning acts according to the output of the Safe Mode Decision module. The proposed system was evaluated through simulations in representative urban settings, demonstrating its effectiveness in handling positioning failures and enhancing localization accuracy by leveraging multiple information sources. 

Orf et al. \cite{orf_modeling_2022} compares pose and time deviations against probability distributions learned during prior error-free mapping runs. Recognizing the position-dependent nature of localization performance, the system employs a grid map, associating each cell with the most suitable probability distribution during the training phase.

A SL assessment method for Kalman Filtering is presented by Griebel et al. \cite{griebel_kalman_2020}. The method uses SL opinions of measurements and a sliding window approach to cope with measurements over time. The opinions are then compared to a reference opinion which represents the assumptions of the Kalman filter. This approach forms the basis for our Localization Diagnosis. But instead of comparison to predefined reference assumptions we compare the behahvior of multiple localization systems to each other.

\section{Subjective Logic}
\label{sec:subjective_logic}
The work presented in this article relies heavily on SL, which was developed by Audun J\o{}sang. SL is thoroughly described in \cite{josang_subjective_2016}, which builds the foundation of this article. In this section we give an overview on SL with focus on the topics necessary for our work. The reader is encouraged to work through \cite{josang_subjective_2016} for more interesting SL subjects. It should be noted that the authors developed a SL library, called \emph{SUBJ}~\cite{orf_subj_2025}, written in C++ and including Python bindings, that implements the SL concepts used in this work. 

SL can be seen as an extension to probabilistic logic, which in turn extends boolean logic to continuous domains, not just \texttt{True}- and \texttt{False}-values. SL also expresses belief in exclusive events as continuous probability values, but also models the uncertainty an entity has about this belief explicitly. By modeling the uncertainty in this way the presence (or abundance) of knowledge can be expressed. While otherwise in classical probability, if an entity is uncertain about the outcome of a particular event often equal probability mass on all events is assumed. This is obviously misleading and undistinguishable of the case that all events are equally likely. SL provides a sound framework with a solid mathematical foundation rendering it valuable for practical applications, especially in cases where information is lacking and must be adequately addressed.

The following section outlines the fundamental principles of SL. It begins with the foundational structure—emph{subjective opinions}—followed by its relationship to the Beta and Dirichlet probability models. Subsequently, key operators relevant to our approach are introduced and discussed.

\subsection{Opinions}
Subjective opinions, short opinions, express a belief with a degree of uncertainty about events or states. These build the domain of the opinion. Belief assigns belief mass to each state, while the uncertainty holds for the entire belief mass distribution. With the additional concept of base rates the prior belief on the general population or background information can be specified. The reasoning entity, that holds an opinion about a subject can also be modeled. This \emph{ownership} underlines the property of SL that multiple reasoning entities can hold different opinions about the same subject. Often it is omitted to explicitly state the owner of an opinion, where it is clear from the context or unnecessary.

\subsubsection{Domains}
The belief of an opinion in SL is defined over a state space $\mathbb{X}$, also called the \emph{domain} of the opinion. The domain $\mathbb{X}$ consists of a set of values, that can be assumed to be states, events or outcomes, e.g. of a random experiment \cite{josang_subjective_2016}. The cardinality $n = |\mathbb{X}|$ or the amount of values in $\mathbb{X}$ is the dimension of the domain. The states of $\mathbb{X}$ are assumed to be exhaustive and exclusive. Thus, the system is only in one state of $\mathbb{X}$ at a time. Also, \cite{josang_subjective_2016} describes hyperdomains, which can be useful e.g. to model systems where multiple states at once are possible. A random variable $X$ over $\mathbb{X}$ takes on values of $\mathbb{X}$, since the states of $\mathbb{X}$ are exclusive.

\subsubsection{Uncertainty Mass}
The \emph{uncertainty} $u_X \in [0,1]$ of an opinion is used to model the presence or abundance of knowledge about the state of the random variable $X$. $u_X$ describes the degree, with which the reasoning entity $A$ is certain or uncertain about the outcome of a random variable $X$ \cite{josang_subjective_2016}. If $A$ is absolutely certain about its belief over $X$ then $u_X = 0$. The corresponding opinion is then called to be \emph{dogmatic}. If, on the other hand, $A$ has no evidence to make any statement about the state of $X$ then $u_X = 1$. In this case the opinion is \emph{vacuous}. 

\subsubsection{Belief Distribution}
An opinion expresses \emph{belief} $\mathbf{b}_X$ of a random variable $X \in \mathbb{X}$ \cite{josang_subjective_2016}. The belief in $X$ corresponds to classic probabilities and indicates the reasoning entities belief of the state the system is in. While in classic probability the additivity requirement is only applied to the belief, in SL it is also extended to the uncertainty. Thus, the belief of a random variable $X \in \mathbb{X}$ is defined as
\begin{align}
    \begin{split}
        \mathbf{b}_X : \mathbb{X} \rightarrow [0,1], \\
        \text{with } u_X + \sum_{x\in\mathbb{X}} \mathbf{b}_X(x) = 1.
    \end{split}
\end{align}

\subsubsection{Base Rate Distribution}
The concept of \emph{base rates} in SL allows to specify a prior belief over the states of $\mathbb{X}$. This prior probability is used to model the a priori knowledge of  possible states the system can be in. A base rate distribution $\mathbf{a}_X$ assigns a prior probability mass to the random variable $X \in \mathbb{X}$. As $\mathbf{a}_x$ is a probability additivity is also required. The base rate distribution is formally defined as
\begin{align}
    \begin{split}
        \mathbf{a}_X : \mathbb{X} \rightarrow [0,1], \\
        \text{with } \sum_{x\in\mathbb{X}} \mathbf{a}_X(x) = 1.
    \end{split}
\end{align}

\subsubsection{Opinion Notation}
Given a random variable $X$ over the states of the domain $\mathbb{X}$, with cardinality $n$, an owner $A$, a belief mass distribution of $\mathbf{b}_X$, uncertainty $u_X$ and base rate distribution $\mathbf{a}_X$ an opinion is notated as
\begin{equation}
\label{eq_multinomial_opinion}
    \omega_{X}^{A} = (\mathbf{b}_X, u_X, \mathbf{a}_X)
\end{equation}

\subsubsection{Opinion Dimension}
An opinion's $\omega_X^A$ dimension is the same as the dimension of its belief vector $\mathbf{b}_X$ and its base rate vector $\mathbf{a}_X$, and thus $n = |\mathbb{X}|$. If $n=2$ the opinion is \emph{binomial}, while for $n>2$ the opinion is \emph{multinomial}. 

A binomial opinion over $Y \in\mathbb{Y}$, with $|\mathbb{Y}| = 2$ defines belief mass only on two states. These states can be explicitly referred to as belief $b_Y$ and disbelief $d_Y$ \cite{josang_subjective_2016}. Here, $d_Y = 1 - b_Y$ applies. Likewise, the binomial opinion's base rate consists also only of two states that sum to $1$, for which usually only the base rate element $a_Y$ corresponding to the belief-state is specified. A binomial opinion is then defined as
\begin{equation}
    \omega_Y^A = (b_Y, d_Y, u_Y, a_Y).
\end{equation}
Whereas multinomial opinions are expressed as in \cref{eq_multinomial_opinion}. In this article we prefer the use of multinomial opinion notation.

\subsection{Beta and Dirichlet Model}
\label{sec:beta_dirichlet_model}
Opinions correspond to classic probabilities \cite{josang_subjective_2016}. Each opinion $\omega_X^A = (\mathbf{b}_X, u_X, \mathbf{a}_X)$ can be projected to a probability by
\begin{align}
    \mathbf{P}_X(x) = \mathbf{b}_X + \mathbf{a}_X u_X, \forall x \in \mathbb{X}. 
\end{align}
The variance of the opinion is then given as
\begin{align}
    \text{Var}_X(x) = \frac{\mathbf{P}_X(x)(1 - \mathbf{P}_X(x))u_x}{W + u_X},
\end{align}
while the constant $W$ is the \emph{non-informative prior weight}, which is usually set to $W = |\mathbb{X}|$ \cite{josang_subjective_2016}.

Opinions use the Dirichlet probability density function (PDF) as its underlying probability model. Each multinomial opinion is equivalent to a Dirichlet PDF over the same domain \cite{josang_subjective_2016}. Usually the Dirichlet PDFs are defined over a strength vector $\alpha_X(x)$. To arrive at a mapping from opinions to Dirichlet PDF, $\alpha_X(x)$ is defined in terms of an evidence vector $\mathbf{r}_X$ and a base rate $\mathbf{a}_X$ \cite{josang_subjective_2016}: 
\begin{align}
    \alpha_X(x) = \mathbf{r}_X(x) + \mathbf{a}_X(x)W, \hspace{0.5em} \text{with } \mathbf{r}_X=x \forall x \in \mathbb{X}.
\end{align}
The Dirichlet PDF over the domain $\mathbb{X}$, with cardinality $n$ in evidence notation $\text{Dir}_X^e(\mathbf{p}_X, \mathbf{r}_X, \mathbf{a}_X)$, with $\mathbf{p}_X$ being the $n$-dimensional random variable, is then, according to \cite{josang_subjective_2016}, defined as
\begin{align}
    \begin{split}
        \text{Dir}_X^e(\mathbf{p}_x, \mathbf{r}_X, \mathbf{a}_X) =\qquad\qquad\qquad& \\ 
        \frac{\Gamma(\sum_{x\in\mathbb{X}}(\mathbf{r}_X(x) + \mathbf{a}_X(x)W))}{\Pi_{x\in\mathbb{X}}(\Gamma(\mathbf{r}_X(x) + \mathbf{a}_X(x)W)}& \\
        \cdot \; \Pi_{x\in\mathbb{X}}\mathbf{p}_X(x)^{\mathbf{r}_X(x)+\mathbf{a}_X(x)W-1)}&,
    \end{split} \\
    \text{where } (\mathbf{r}_X(x) + \mathbf{a}_X(x)W) \geq 0. \nonumber
\end{align}

The mapping of Dirichlet PDFs and multinomial opinions is described in \cite{josang_subjective_2016} as follows. Given a Dirichlet PDF $\text{Dir}_X^e(\mathbf{p}_X, \mathbf{r}_X, \mathbf{a}_X)$ the parameters of a multinomial opinion $\omega_X = (\mathbf{b}_X, u_X, \mathbf{a}_X)$ are then
\begin{align}
\label{eq:dirichlet_opinion_mapping}
    \begin{split}
        \mathbf{b}_X(x) &= \frac{\mathbf{r}_X(x)}{W + \sum_{x_i \in \mathbb{X}}\mathbf{r}_X(x_i)}, \\
        u_X &= \frac{W}{W+\sum_{x_i \in \mathbb{X}}\mathbf{r}_X(x_)}.
    \end{split}
\end{align}
Vice versa the mapping of a given opinion $\omega_X = (\mathbf{b}_X, u_X, \mathbf{a}_X)$ to a Dirichlet PDF $\text{Dir}_X^e(\mathbf{p}_X, \mathbf{r}_X, \mathbf{a}_X)$ is given by
\begin{equation}
\begin{alignedat}{2}
    &\text{For }u_X \neq 0: \quad &&\text{For }u_X = 0: \\ 
    \hline \\
    &\mathbf{r}_X(x) = \frac{W\mathbf{b}_X(x)}{u_X}, \quad &&\mathbf{r}_X(x) = \mathbf{b}_X(x) \cdot \infty, \\
    &1 = u_X + \sum_{x_i \in \mathbb{X}}\mathbf{b}_X(x_i) \quad &&1 = \sum_{x_i \in \mathbb{X}}\mathbf{b}_X(x_i).
\end{alignedat}
\end{equation}
For the special case of a binomial opinion the corresponding Dirichlet PDF is also $2$-dimensional and thus a Beta PDF.

\subsection{Operators}
\label{sec:operators}
The power of SL comes with operators defined on opinions. Numerous operators are defined in \cite{josang_subjective_2016}, from which we only use a few in this article. We briefly describe the used operators in the following. The reader is refered to \cite{josang_subjective_2016} for in-depth information on SL and operators.

\subsubsection{Degree of Conflict}
\label{sec:degree_of_conflict}
Disagreeing or conflicting opinions are frequently encountered in SL, especially when different owners express opinions on the same subject. Measuring these conflicts can be done with the \emph{Degree of Conflict} (DC) operator. Given two opinions $\omega_X^A$ and $\omega_X^B$ on the variable $X$, the degree of conflict is defined as
\begin{align}
    \begin{split}
        \text{DC}(\omega_X^A, \omega_X^B) =\qquad\qquad\qquad& \\
        \frac{\sum_{x\in\mathbb{X}} | \mathbf{P}_X^A(x) - \mathbf{P}_X^B(x)}{2} &- (1-u_X^A)(1-u_X^B).
    \end{split}
\end{align}

\subsubsection{Cumulative Belief Fusion}
\label{sec:cumulative_belief_fusion}
Fusing two opinions $\omega_X^A$ and $\omega_X^B$ together to form a new opinion $\omega_X^{(A \diamond B)}$ that combines both assumptions can be fulfilled with the \emph{Aleatory Cumulative Fusion} (CF) operator \cite{josang_subjective_2016}. 
\begin{align}
\label{eq:cumulative_belief_fusion}
    \omega_X^{(A \diamond B)} = \omega_X^A \oplus \omega_X^B
\end{align}
The $\oplus$-sign is used in this article to denote CF. We omit repeating the calculations of the CF operator here and refer the reader to \cite{josang_subjective_2016}. By fusing opinions with contrary beliefs and different uncertainty, the opinion with less uncertainty is dominant in the resulting opinion. 

\subsubsection{Cumulative Unfusion}
\label{sec:cumulative_unfusion}
Similar to fusing two opinions an operator for unfusion exists \cite{josang_subjective_2016}. For a given opinion $\omega_X^{(A \diamond B)}$, which originated from fusion of two opinions, where one opinion $\omega_X^B$ is known, the unfusion is
\begin{align}
\label{eq:cumulative_unfusion}
    \omega_X^A = \omega_X^{(A \diamond B)} \ominus \omega_X^B
\end{align}
We use the $\ominus$ operator to eliminate a single measurement from a fused opinion of multiple measurement opinions. Again, we omit the calculations of the cumulative unfusion and refer the reader to \cite{josang_subjective_2016}. 

\subsubsection{Trust Discount}
\label{sec:trust_discount}
Similar to \cite{griebel_self-assessment_2022} we also employ the trust discount operator to reflect the decay of information over time. With a given trust discount probability $p_{td}$ the trust discounted opinion $\omega_X^{[A; p_{td}]} = \text{TD}(\omega_X^A, p_{td})$ on an opinion $\omega_X^A$ is given by
\begin{align}
    \mathbf{b}_X^{[A;p_{td}]}(x) &= p_{td}\mathbf{b}_x^A(x), \\
    u_X^{[A;p_{td}]} &= 1 - p_{td} \sum_{x\in\mathbb{X}}\mathbf{b}_X^A(x), \\
    \mathbf{a}_X^{[A;p_{td}]}(x) &= \mathbf{a}_X^A(x).
\end{align}

\subsubsection{Normal Multiplication}
\label{sec:normal_multiplication}
Multiplication of opinions on different random variables is necessary for computing an opinion on the joint variable. Consider an opinion $\omega_X$ on the variable $X \in \mathbb{X}$ with cardinality $n$ and an opinion $\omega_Y$ on $Y\in\mathbb{Y}$ with cardinality $k$. The resulting joint opinion $\omega_{XY}$ is then defined on $\mathbb{X} \times \mathbb{Y}$ with cardinality $n\cdot k$. Normal multiplication of two opinions can be conducted if both variables $X$ and $Y$ are independent of each other. Again, we omit the extensive formulas for the normal multiplication and refer the reader to \cite{josang_subjective_2016}.

\section{Multi-Source Localization Performance Assessment}

In AD multiple localization systems are present. This helps to compensate for failures and also increases accuracy. Our method for assessing localization performance takes advantage of this by identifying discrepancies in each and between all localizations. Failures in localization systems are in general a deviation or discrepancy between the real (ground truth) position and the measured position. While every localization suffers from sensor noise as well as model assumptions in the method itself the application dictates the required localization accuracy. Especially in AD with its safety requirements centimeter-level accuracy is essential. Even more, as most AD software components rely on the vehicle's localization special emphasis must be placed on its robustness and reliability.

We assume each localization system updates its position estimate at discrete points in time. While typically a localization outputs a globally referenced position, we focus on the relative position update. Hence, vehicle odometry can also be considered, which improves overall robustness. Typical failures in localization systems can also be distinguished between incorrect absolute/global and relative positions. We assume that a correct initial position estimate is given and recognize failures on-line during localization updates. Such failures include drifts, jumps and stops. The occurrence of these failures depends on the localization method.

\subsection{Localization Input Opinion Generation}
\label{sec:opinion_representation}
The global position inputs from a localization system $L$ at a given time point $t$ are $x_t^L$ and $y_t^L$. Since the global position is not informative for detecting the aforementioned faults for some localization methods (e.g. odometry), we consider the relative position $\Delta x_t^{L} = x_t^L - x_{t-1}^L$. 

In the SL based approach, opinions need to be derived that represent the localization inputs. As opinions have discrete states, a histogram mapping is employed. We utilize two histograms $H_x^L \in \mathbb{R}^{n}$ with $n$ discrete bins and $H_y^L \in \mathbb{R}^{m}$ with $m$ bins for each input dimension $x$ and $y$. The histograms are defined with evenly spaced bins for a range of input values $[x_{\text{min}}, x_{\text{max}}]$ and $[y_{\text{min}}, y_{\text{max}}]$. The bin borders of $H_x^L$ being $[x_i, x_{i+1}]$ with $x_i = x_{\text{min}} + i \cdot w_x$, where $w_x = \frac{x_{\text{max}} - x_{\text{min}}}{n}$ for $i \in \{1,\dots,n-1\}$. Respectively the bin borders of $H_y^L$ are $[y_i, y_{i+1}]$ with $y_i = y_{\text{min}} + i \cdot w_y$, where $w_y = \frac{y_{\text{max}} - y_{\text{min}}}{m}$ for $i \in \{1,\dots,m-1\}$. Additionally, $x_0 = y_0 = -\infty$ and $x_n = x_m = \infty$. Values $\Delta x_t^L$ and $\Delta y_t^L$ are then assigned to a bin with $\text{bin}_x(\Delta x_t^L) \in \{0,\dots, n\}$ and $\text{bin}_y(\Delta y_t^L) \in \{0,\dots, m\}$. The histograms for a single time point $t$ are then formed by $H_{t,x}^L(\text{bin}_x(\Delta x_t^L)) = H_{t,y}^L(\text{bin}_y(\Delta y_t^L)) = 1$. The histograms are created from scratch for every time point. Since considering only a single input, $H_{t,x}^L$ and $H_{t,y}^L$ are normalized histogram.
Localization systems used in this article are a SLAM based localization $S$, GNSS based localization $G$ and the odometry of the vehicle $O$. Note that the odometry outputs relative positions and thus its values are used as is for building the histograms $H_{t,x}^O$ and $H_{t,y}^O$.

Mapping the input values of a localization system $L$ to an opinion is straightforward with the evidence notation of Dirichlet PDFs (\cref{sec:beta_dirichlet_model}). By using the histograms of the input values at a distinct time point $t$ as the evidence vector $\bm{r}_{X} = H_{t,x}^L$ and $\bm{r}_{Y} = H_{t,y}^L$ for the random variables $X \in \mathbb{X}$ with $|\mathbb{X}| = n$ and $Y \in \mathbb{Y}$ with $|\mathbb{Y}| = m$, the opinions $\omega_{t,X}^L = (\bm{b}_{t,X}^L, u_{t,X}^L, \bm{a}_X^L)$ and $\omega_{t,Y}^L = (\bm{b}_{t,Y}^L, u_{t,Y}^L, \bm{a}_Y^L)$ are calculated with \cref{eq:dirichlet_opinion_mapping} for given base rates $\bm{a}_X^L$ and $\bm{a}_Y^L$. For simplicity we assume $X$ and $Y$ to be independent. Then, to arrive at a single input opinion $\omega_{t,Z}^L = \omega_{t,X}^L \cdot \omega_{t,Y}^L$ is calculated with the normal multiplication operator (see \cref{sec:normal_multiplication}), by assuming independence, where $Z = XY$ is the joint random variable with $|Z| = k = n\cdot m$. This opinion then represents the input from a single localization system $L$ at a distinct time point $t$. Since $Z$ is equally defined for all localization systems, we omit stating it explicitly in the remainder of this article and only write $\omega_t^L$.

\subsection{Sliding-Window Approach}
The goal is to model all localization systems' behavior over time and compare them to each other. As each localization system has unique advantages and disadvantages the assumption is that most failures can be recognized in this way. An input of a localization system $L$ at time point $t$ is $\omega_t^L$ (see \cref{sec:opinion_representation}). Comparing such single inputs is less helpful because depending on the localization systems the differences would almost always be high. Instead we adopt the SL based sliding-window approach from \cite{griebel_self-assessment_2022}. There, two sliding windows are used for different time spans to focus on sudden changes while at the same time having a robust behavior model. In the following we adopt this approach to our localization scenario.

A short term (ST) window $\omega_{\text{st}}^L$ is built by fusing the measurement opinion $\omega_t^L$ into it (see \cref{sec:cumulative_belief_fusion}). The opinion $\omega_{\text{st}}^L$ then represents all measurements of $L$ for a given time frame. The amount of fused measurements is $| \omega_{\text{st}}^L |$. If $l_{\text{st}}$ measurements are fused into $\omega_{\text{st}}^L$ the oldest element $\omega_{\text{old}}^L = \omega_{t-l_{\text{st}}}^L$ is removed by unfusion (see \cref{sec:cumulative_unfusion}. For each time point $t$ the ST window opinion $\bar{\omega}_{\text{st}}^L$ from the previous iteration is updated by calculating
\begin{align}
    \omega_{\text{st}}^L = \left\{ \begin{aligned}
        (&\bar{\omega}_{\text{st}}^L \oplus \omega_t^L) \ominus \omega_{\text{old}}^L \quad &&\text{if } |\bar{\omega}_{\text{st}}^L | \geq l_{st}^L \\
        &\bar{\omega}_{\text{st}}^L \oplus \omega_t^L \quad &&\text{if } |\bar{\omega}_{\text{st}}^L | < l_{\text{st}}^L.
    \end{aligned} \right.
\end{align}

The oldest element $\omega_{\text{old}}^L$ which is unfused from $\omega_{\text{st}}^L$ is then added to the long term (LT) window $\omega_{\text{lt}}^L$ also by aleatory cumulative fusion (see \cref{sec:cumulative_belief_fusion}). While the LT window is used for a longer baseline of the localization system's behavior, very old measurements shouldn't influence the model much. Thus, trust discount (see \cref{sec:trust_discount}) with a given probability $p_{td}$ is applied to the LT window in each iteration. This equals to weighing the belief of each measurement $\omega_{t-l_{\text{st}}-x}^L$ which was added to the LT window $x$ time points ago with a weight of $p_{td}^{x}$. At a given time point $t$ the LT window $\omega_{\text{lt}}^L$ is then calculated with the LT window from the previous iteration $\bar{\omega}_{\text{lt}}^L$ and the oldest element $\omega_{\text{old}}^L$, which was unfused from $\omega_{\text{st}}^L$ by applying
\begin{align}
    \omega_{\text{lt}}^L = \text{TD}(\bar{\omega}_{lt}^A, p_{td}) \oplus \omega_{\text{old}}^L.
\end{align}

With the ST and LT windows $\omega_{\text{st}}^L$ and $\omega_{\text{lt}}^L$ an opinion per iteration can be generated with which the localization system's can be compared. This is done by fusing $\omega_{\text{st}}^L$ and $\omega_{\text{lt}}^L$ with aleatory cumulative belief fusion (see \cref{sec:cumulative_belief_fusion}). But to account for sudden changes during localization the fusion is only performed when $\omega_{\text{st}}^L$ and $\omega_{\text{lt}}^L$ are similar. This is achieved with the DC operator (see \cref{sec:degree_of_conflict}) with a given threshold $\theta^L$. An opinion $\hat{\omega}_t^L$ that reflects the behavior of $L$ at time point $t$ is then calculated with
\begin{align}
    \hat{\omega}_{t}^L = \left\{ \begin{aligned}
        &\omega_{\text{st}}^L \quad &&\text{if } \text{DC}(\omega_{\text{st}}^L, \omega_{\text{lt}}^L) > \theta^L \\
        &\omega_{\text{st}}^L \oplus \omega_{\text{lt}}^L \quad &&\text{if } \text{DC}(\omega_{\text{st}}^L, \omega_{\text{lt}}^L) \leq \theta^L.
    \end{aligned} \right.
\end{align}

In each time point $t$ the opinion $\hat{\omega}_t^L$ is then used for comparison. For a given localization system $L$ with its opinion $\hat{\omega}_t^L$ and a reference localization system $L'$ with the reference opinion $\hat{\omega}_t^{L'}$ the comparison yields a measure for difference $\delta_t^{LL'}$. Additionally, an uncertainty measure $u_t^{LL'}$ is generated by obtaining the uncertainty $u_t^L$ of $\hat{\omega}_t^L$. The difference measure $\delta_t^{LL'}$ together with the uncertainty $u_t^{LL'}$ is calculated with the DC operator (see \cref{sec:degree_of_conflict}) by
\begin{align}
    \delta_t^{LL'} &= \text{DC}(\hat{\omega}_t^L,\hat{\omega}_t^{L'})\\
    u_t^{LL'} &= u_t^L(\hat{\omega}_t^{L}).
\end{align}
By using this measure the difference of localization systems can be expressed with an explicit statement on the uncertainty. Note, that $\delta_t^{LL'} \neq \delta_t^{L'L}$ and $u_t^{LL'} \neq u_t^{L'L}$.

\section{Evaluation}

\begin{figure}
    \begin{subfigure}[ht]{\columnwidth}
		\includesvg[width=\textwidth]{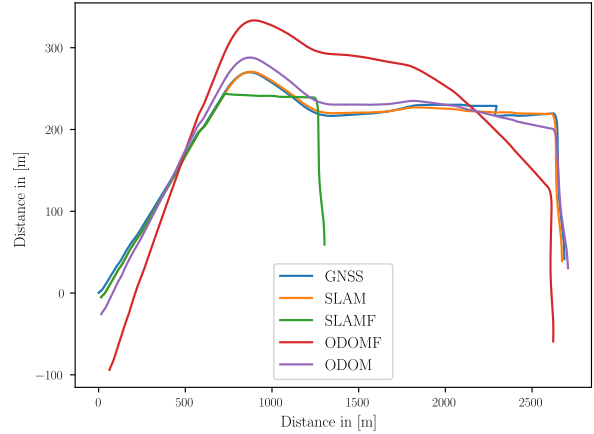}
    \end{subfigure}\hfill
    \caption{The input data for the self-assessment comprises five distinct trajectories: two odometry-based trajectories utilizing different motion models, two SLAM-derived trajectories —~one of which includes error injection — and a single trajectory obtained from GNSS measurements.}
    \label{fig:input_data}
\end{figure}

\begin{figure*}[ht]
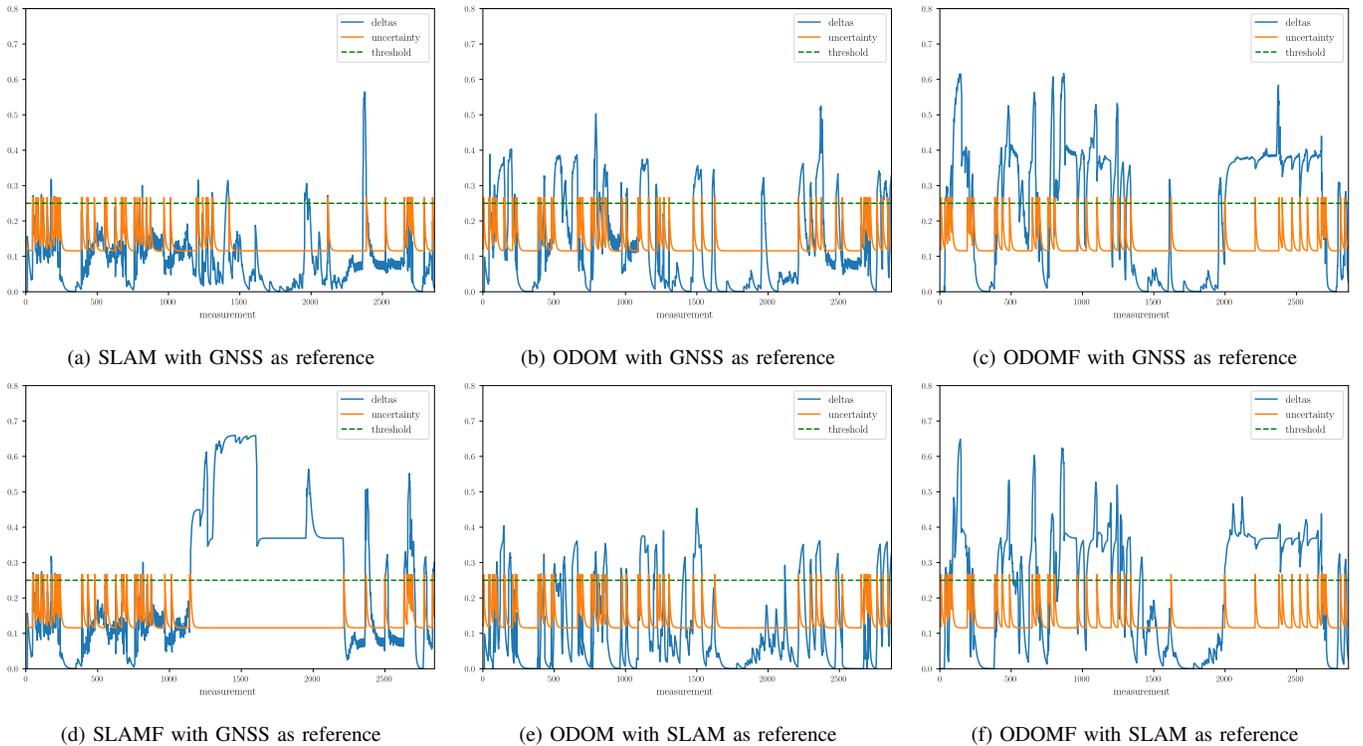
  
     \begin{subfigure}[b]{0.33\textwidth}
		\includesvg[width=\textwidth]{resources/sa_SLAM_with_GNSS_ref.csv.svg}
        \caption{SLAM with GNSS as reference}
        \label{fig:SLAM-GNSS}
    \end{subfigure}\hfill
    \begin{subfigure}[b]{0.33\textwidth}
		\includesvg[width=\textwidth]{resources/sa_ODOM_with_GNSS_ref.csv.svg}
        \caption{ODOM with GNSS as reference}
        \label{fig:ODOM-GNSS}
    \end{subfigure}\hfill
    \begin{subfigure}[b]{0.33\textwidth}
		\includesvg[width=\textwidth]{resources/sa_ODOMF_with_GNSS_ref.csv.svg}
        \caption{ODOMF with GNSS as reference}
        \label{fig:ODOMF-GNSS}
    \end{subfigure}\hfill
    \begin{subfigure}[b]{0.33\textwidth}
		\includesvg[width=\textwidth]{resources/sa_SLAMF_with_GNSS_ref.csv.svg}
        \caption{SLAMF with GNSS as reference}
        \label{fig:SLAMF-GNSS}
    \end{subfigure}\hfill
    \begin{subfigure}[b]{0.33\textwidth}
		\includesvg[width=\textwidth]{resources/sa_ODOM_with_SLAM_ref.csv.svg}
        \caption{ODOM with SLAM as reference}
        \label{fig:ODOM-SLAM}
    \end{subfigure}\hfill
    \begin{subfigure}[b]{0.33\textwidth}
		\includesvg[width=\textwidth]{resources/sa_ODOMF_with_SLAM_ref.csv.svg}
        \caption{ODOMF with SLAM as reference}
        \label{fig:ODOMF-SLAM}
    \end{subfigure}\hfill
    \caption{Cross-validation results for six examples are presented. The blue line indicates the changes ($\delta_t^{LL'}$) in the Degree of Conflict, the green line represents the threshold $\theta^L$ for event triggering, and the orange line depicts the confidence level of the opinion ($u_t^L$). The figure demonstrates the effectiveness of our approach in detecting both discrete anomalies — such as the jump observed in SLAM with GNSS (\cref{fig:SLAM-GNSS}) as reference around time point 2400 — and continuous errors, as seen in SLAMF with GNSS as reference, \cref{fig:SLAMF-GNSS}. Additionally, the method allows for evaluation of odometry model performance, as reflected by the differing magnitudes between ODOM and ODOMF results.}
    \label{fig:self-assessment-result}
\end{figure*}

In this section, we present the evaluation of our localization approach using real-world data provided by the CoCarNextGen \cite{heinrich_cocar_nodate}. We compare five distinct localization methods: the built-in GNSS, wheel-based odometry, and the Chefs-Kiss SLAM algorithm (SLAM) \cite{ochs_chefs_nodate}. Each method exhibits unique strengths and limitations, making it well-suited for our scenario's comparative analysis. To showcase the capabilities of our approach, we additionally inject an error in the Chefs-KISS (SLAMF), by keeping the position after a tunnel entry constant. And for the odometry, we utilize a different model for the odometry estimation from the raw data, which does not approximate the testing vehicle very well (ODOMF).

The evaluation was conducted during a drive through a tunnel in Karlsruhe, Germany. \Cref{fig:input_data} provides an overview of the input data. The vehicle enters the tunnel at approximately time point 1000 and exits at 2400. This specific environment presents a challenging use case for localization due to the temporary loss of GNSS signal reception inside the tunnel.

The GNSS system is integrated into CoCar NextGen and includes an Inertial Measurement Unit for dead-reckoning during GNSS outages. \Cref{fig:input_data} shows that a significant discontinuity occurs at the tunnel exit due to accumulated drift during GNSS signal loss and subsequent correction once signal acquisition resumes. 
In contrast, the wheel odometry provides consistent and smooth localization throughout the tunnel segment. However, it is susceptible to cumulative drift, particularly during longer trajectories or under varying road surface conditions. 
The Chefs-Kiss SLAM method consistently delivers reliable localization across the entire sequence. In addition, to show that SLAM can also be diagnosed, we introduce an error injection for evaluation, as seen in \Cref{fig:input_data}.

The preprocessing step for the self-assessment involves downsampling all localization data to a common temporal resolution. This unified time delta is determined by the slowest of the evaluated localization methods. In our case, the limiting factor is the SLAM-based approach, which operates at the frame rate of the employed LiDAR sensors. The LiDAR sensor used in the experiments captures environmental data at a frequency of \SI{10}{\hertz}. To ensure consistency across all methods, the localization data is resampled accordingly. Additionally, the movement of the vehicle is inspected for each spatial dimension, specifically the longitudinal (x-axis) and lateral (y-axis) direction. The input data is processed as described in \cref{sec:opinion_representation}.

\subsection{Results}

\begin{figure}
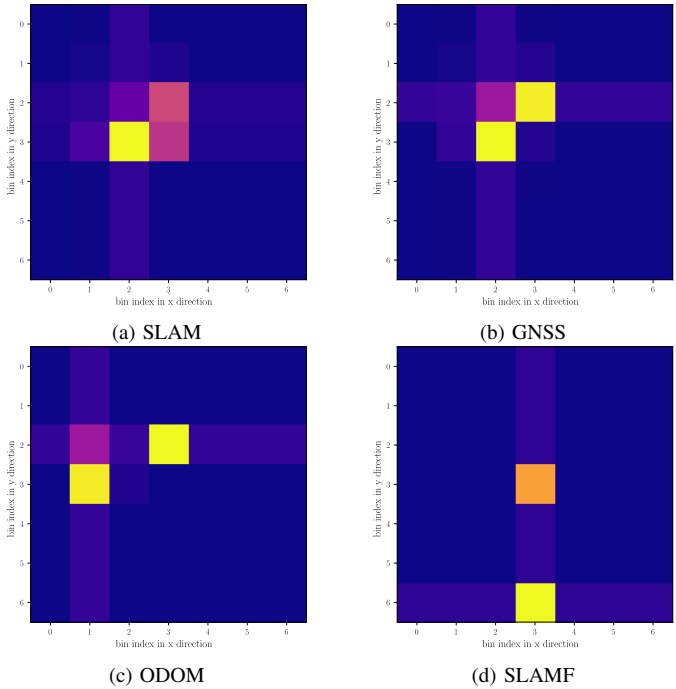

     \begin{subfigure}[t]{0.45\columnwidth}
		\includesvg[width=\textwidth]{resources/1500_ref_SLAM_heatmap.svg}
        \caption{SLAM}
        \label{fig:input_ODOM_1500}
    \end{subfigure}\hfill
     \begin{subfigure}[t]{0.45\columnwidth}
		\includesvg[width=\textwidth]{resources/1500_ref_GNSS_heatmap.svg}
        \caption{GNSS}
        \label{fig:input_ref_GNSS_1500}
    \end{subfigure}\hfill
     \begin{subfigure}[t]{0.45\columnwidth}
		\includesvg[width=\textwidth]{resources/1500_ref_ODOM_heatmap.svg}
        \caption{ODOM}
        \label{fig:input_ODOM_2387}
    \end{subfigure}\hfill
     \begin{subfigure}[t]{0.45\columnwidth}
		\includesvg[width=\textwidth]{resources/1500_sla_SLAMF_heatmap.svg}
        \caption{SLAMF}
        \label{fig:slamf_1500}
    \end{subfigure}\hfill
    \caption{Input data for SLAM, GNSS, ODOM, and SLAMF at time step 1500 are shown. The first row exhibits similar spatial distributions across the modalities, resulting in a low degree of conflict. In contrast, the second row reveals more pronounced discrepancies, which correspond to a significant peak in the degree of conflict observed in \cref{fig:self-assessment-result}.}
    \label{fig:opinion}
\end{figure}

For evaluation, we employ cross-validation between pairs of localization inputs. \Cref{fig:self-assessment-result} presents six representative examples of these cross-validations. Given the near-cumulative nature of the trust-based assessment, we investigated the following combinations: SLAM–GNSS, ODOM–GNSS, ODOMF–GNSS, SLAMF–GNSS, ODOM–SLAM, and ODOMF–SLAM.

As a first step, we present the output opinion $\hat{\omega}_t^L$ from various localization methods at different time points, as shown in \cref{fig:opinion}. The first row displays the current input data from SLAM in combination with GNSS. It consists of a pair of heatmaps: the right side shows the current input distribution, while the left side displays the corresponding reference opinion. Both heatmaps visualize the distribution within the 2D histogram defined in \cref{sec:opinion_representation}. In the first example, no noticeable difference between the two distributions is visible, resulting in a low DC, which is also reflected in \cref{fig:SLAM-GNSS}.

The second row, however, depicts two instances of error. In these cases, discrepancies arise between the odometry and SLAM or GNSS alignments, yet a perceptible correlation remains. As a result, the DC value exceeds the threshold, although the amplitude remains low. In contrast, SLAMF exhibits a higher amplitude than ODOM, as evident in \cref{fig:slamf_1500}, which also corresponds to an elevated DC value.

In the first combination (SLAM–GNSS), a distinct spike is observed around timestep 2400, as shown in \cref{fig:SLAM-GNSS}. This corresponds exactly to the moment when the GNSS system resumes receiving correction data, resulting in a sudden positional jump, as depicted in \cref{fig:input_data}. This scenario represents a primary use case for our localization framework: detecting abrupt, discrete changes in localization estimates.

The second use case involves identifying prolonged disturbances in localization performance. This is exemplified in \cref{fig:SLAMF-GNSS}, where the SLAM algorithm is deliberately injected with a static position. This results in a persistent error, clearly visible as a continuous spike from timestep 1100 to 2250—coinciding with the duration of the injected error.

Cross-validation involving odometry-based localization reveals a consistent pattern: odometry performs poorly as a standalone long-term localization method, with defect detections occurring almost continuously. This issue is amplified in the ODOMF variant, where the residual errors are even more pronounced. However, as shown in \cref{fig:ODOM-GNSS} and \cref{fig:ODOMF-GNSS}, the period between timesteps 1300 and 2500 shows fewer anomalies, coinciding with the activation of GNSS dead reckoning. During this interval, the similarity between odometry-based and IMU-integrated estimates leads to lower cross-validation discrepancies.

In contrast, comparisons between odometry and SLAM reveal minor but persistent deviations, with the exception of a short interval between timesteps 1500 and 2000, during which the results align well. Notably, as illustrated in \cref{fig:ODOM-SLAM} and \cref{fig:ODOMF-SLAM}, the fine-tuned odometry model ODOM exhibits improved performance, reducing discrepancies relative to standard odometry ODOMF.

\section{Conclusion}

In this work, we presented a new approach for multi-modal localization validation. it relies as a backbone on SL and cross-validation of different localization methods. The evaluation is conducted with real world data which were extended with error injection to highlight all capabilities of the framework. It can detect short spikes, persistent errors and low quality models.

\section*{Acknowledgment}
This paper was created in the "Country 2 City - Bridge" project of the "German Center for Future Mobility", which is funded by the German Federal Ministry for Digital and Transport.

\bibliographystyle{splncs04}
\bibliography{references_so}

\begin{thebibliography}{10}
\providecommand{\url}[1]{\texttt{#1}}
\providecommand{\urlprefix}{URL }
\providecommand{\doi}[1]{https://doi.org/#1}

\bibitem{delmotte_detection_2008}
Delmotte, F., Gacquer, G.: Detection of defective sources with belief functions. In: Proceedings of 12th {International} {Conference} on {Information} {Processing} and {Management} of {Uncertainty} in {Knowledge}-{Based} {Systems}, {Malaga}, {Spain}. vol.~2227, p. 337344 (2008)

\bibitem{dosovitskiy_carla_2017}
Dosovitskiy, A., Ros, G., Codevilla, F., Lopez, A., Koltun, V.: {CARLA}: {An} {Open} {Urban} {Driving} {Simulator}. In: Proceedings of the 1st {Annual} {Conference} on {Robot} {Learning}. pp. 1--16 (2017)

\bibitem{griebel_kalman_2020}
Griebel, T., Müller, J., Buchholz, M., Dietmayer, K.: Kalman {Filter} {Meets} {Subjective} {Logic}: {A} {Self}-{Assessing} {Kalman} {Filter} {Using} {Subjective} {Logic}. In: 2020 {IEEE} 23rd {International} {Conference} on {Information} {Fusion} ({FUSION}). pp.~1--8 (Jul 2020)

\bibitem{griebel_self-assessment_2022}
Griebel, T., Müller, J., Geisler, P., Hermann, C., Herrmann, M., Buchholz, M., Dietmayer, K.: Self-{Assessment} for {Single}-{Object} {Tracking} in {Clutter} {Using} {Subjective} {Logic}. In: 2022 25th {International} {Conference} on {Information} {Fusion} ({FUSION}). pp.~1--8 (Jul 2022)

\bibitem{heinrich_cocar_nodate}
Heinrich, M., Zipfl, M., Uecker, M., Ochs, S., Gontscharow, M., Fleck, T., Doll, J., Schörner, P., Hubschneider, C., Zofka, M.R., Viehl, A., Zöllner, J.M.: {CoCar} {NextGen}: a {Multi}-{Purpose} {Platform} for {Connected} {Autonomous} {Driving} {Research}

\bibitem{josang_subjective_2016}
Jøsang, A.: Subjective logic, vol.~3. Springer (2016)

\bibitem{ochs_chefs_nodate}
Ochs, S., Heinrich, M., Schörner, P., Zofka, M.R., Zöllner, J.M.: A {Chefs} {KISS} -- {Utilizing} semantic information in both {ICP} and {SLAM} framework, arXiv:2504.02086

\bibitem{orf_subj_2025}
Orf, S.: {SUBJ} - {Subjective} {Logic} {Library} (2025), \url{https://github.com/fzi-forschungszentrum-informatik/SUBJ/}

\bibitem{orf_modeling_2022}
Orf, S., Lambing, N., Ochs, S., Zofka, M.R., Zöllner, J.M.: Modeling {Localization} {Uncertainty} for {Enhanced} {Robustness} of {Automated} {Vehicles}. In: 2022 {IEEE} 18th {International} {Conference} on {Intelligent} {Computer} {Communication} and {Processing} ({ICCP}). pp. 175--182 (Sep 2022), iSSN: 2766-8495

\bibitem{rodriguez-arozamena_fail-safe_2023}
Rodríguez-Arozamena, M., Aranguren-Mendieta, I., Pérez, J., Zubizarreta, A.: Fail-{Safe} {Decision} {Architecture} for {Positioning} {Failures} on {Automated} {Vehicles}. In: 2023 {IEEE} {Smart} {World} {Congress} ({SWC}). pp.~1--8 (Aug 2023)

\bibitem{viana_reconfigurable_2022}
Viana, K., Zubizarreta, A., Diez, M.: A {Reconfigurable} {Framework} for {Vehicle} {Localization} in {Urban} {Areas}. Sensors  \textbf{22}(7), ~2595 (Jan 2022)

\end{thebibliography}

\end{document}